\documentclass[10pt,twocolumn,letterpaper]{article}

\usepackage{iccv}
\usepackage{times}
\usepackage{epsfig}
\usepackage{graphicx}
\usepackage{amsmath}
\usepackage{amssymb}
\usepackage{multicol}
\usepackage{multirow}
\usepackage{epsfig}
\usepackage{booktabs}

\usepackage[pagebackref=true,breaklinks=true,letterpaper=true,colorlinks,bookmarks=false]{hyperref}

\iccvfinalcopy 

\ificcvfinal\fi

\begin{document}

\title{Learning a More Continuous Zero Level Set in Unsigned Distance Fields through Level Set Projection}

\author{Junsheng Zhou\textsuperscript{1}\thanks{Equal contribution.},\hspace {1mm}  Baorui Ma$^{1,2,*}$, Shujuan Li$^1$, Yu-Shen Liu$^1$\thanks{The corresponding author is Yu-Shen Liu.}, Zhizhong Han$^3$\\
School of Software, Tsinghua University, Beijing, China$^1$\\
Beijing Academy of Artificial Intelligence, Beijing, China$^2$\\
Department of Computer Science, Wayne State University, Detroit, USA$^3$\\
{\tt\small zhoujs21@mails.tsinghua.edu.cn,\hspace {3mm}brma@baai.ac.cn,\hspace {3mm}lisj22@mails.tsinghua.edu.cn}\\
{\tt\small liuyushen@tsinghua.edu.cn,\hspace {3mm}h312h@wayne.edu}
}

\maketitle
\ificcvfinal\thispagestyle{empty}\fi

\begin{abstract}

Latest methods represent shapes with open surfaces using unsigned distance functions (UDFs). They train neural networks to learn UDFs and reconstruct surfaces with the gradients around the zero level set of the UDF. However, the differential networks struggle from learning the zero level set where the UDF is not differentiable, which leads to large errors on unsigned distances and gradients around the zero level set, resulting in highly fragmented and discontinuous surfaces. To resolve this problem, we propose to learn a more continuous zero level set in UDFs with level set projections. Our insight is to guide the learning of zero level set using the rest non-zero level sets via a projection procedure. Our idea is inspired from the observations that the non-zero level sets are much smoother and more continuous than the zero level set. We pull the non-zero level sets onto the zero level set with gradient constraints which align gradients over different level sets and correct unsigned distance errors on the zero level set, leading to a smoother and more continuous unsigned distance field. 
We conduct comprehensive experiments in surface reconstruction for point clouds, real scans or depth maps, and further explore the performance in unsupervised point cloud upsampling and unsupervised point normal estimation with the learned UDF, which demonstrate our non-trivial improvements over the state-of-the-art methods. Code is available at \url{https://github.com/junshengzhou/LevelSetUDF}.

\end{abstract}

\section{Introduction}
Surface reconstruction from point clouds plays an important role in vision, robotics and graphics. Recently, neural implicit functions \cite{park2019deepsdf, mildenhall2020nerf, mescheder2019occupancy} have achieved remarkable results by representing 3D geometries with signed distance functions (SDFs) \cite{ma2021neural, chabra2020deep, wang2021neus}, occupancy \cite{mescheder2019occupancy, peng2020convolutional, oechsle2021unisurf} or unsigned distance functions (UDFs) \cite{Zhou2022CAP-UDF, chibane2020neural, guillard2021meshudf}. SDFs and occupancy explore the internal-to-external relations, which are limited to reconstruct closed surfaces. Researchers therefore turn to UDFs, which is a more general function, to represent shapes with arbitrary typologies. 

Learning an accurate zero level set of UDF is the key factor for representing complex geometries and supporting 3D applications \cite{On-SurfacePriors, li2019pu, guerrero2018pcpnet, chen2023unsupervised}. However, the differential neural networks struggle from learning the zero level set where the UDF is not differentiable. This leads to unstable optimizations, which results in large errors on unsigned distances and gradients around the zero level set, causing failures in downstream 3D applications. Taking surface reconstruction for example, SOTA methods \cite{Zhou2022CAP-UDF, guillard2021meshudf} extract surfaces from UDFs by judging gradient directions around the zero level set, where the unreliable gradients lead to highly fragmented and discontinuous surfaces. 

To resolve this issue, we introduce two novel constraints on the zero level set in forms of losses for learning accurate UDF without ground truth unsigned distances. The key insight is from the observation that despite the unreliable zero level set, the non-zero level sets learned by UDF are quite convincing, which can serve as an accurate guidance for the optimization on zero level set. We therefore propose the level set projection constraint to achieve a smoother and more continuous zero level set of UDF by enforcing the parallelism between the gradient of a query in a non-zero level set and the gradient of its projection to the zero level set. Moreover, we introduce an adaptive weighting strategy to force the neural network to focus more on the optimization nearer to the zero level set. 

The level set projection constraint aims to guide the gradients at zero level set, where we propose another unsigned distance loss as a supplement for correcting distance errors at the zero level set. 
To bring a more robust guidance to the optimization of zero level set, we introduce an addition gradient constraint on queries to learn more accurate and regular non-zero level sets of UDF by encouraging the gradient orthogonal to the tangent plane of the surface.

Based on our learned stable UDF with an accurate zero level set, we explore three different 3D applications, including surface reconstruction, unsupervised point normal estimation and unsupervised point cloud upsampling. Our improvements over unsupervised or even supervised baselines justify both our effectiveness and the importance of an accurate and continuous zero level set to the representation ability of neural unsigned fields.

Our main contributions can be summarized as follows.
\begin{itemize}
    \item We propose two novel constraints on UDF to achieve a smoother and more continuous zero level set for learning to represent shapes or scenes without ground truth unsigned distances.
    \item We justify that an accurate and continuous zero level set is the key factor to represent complex 3D geometries and supporting 3D downstream applications.
    \item We explore three different 3D applications with our learned UDF, including surface reconstruction, unsupervised point normal estimation and unsupervised point cloud upsampling, where we show our superiority over the state-of-the-art supervised / unsupervised methods.
\end{itemize}

\section{Related Work}
With the rapid development of deep learning, the neural networks have shown great potential in 3D applications \cite{qi2017pointnet, han2020shapecaptioner, xiang2022snowflake, wen2022pmp, xiang2021snowflakenet, wen2021pmp, zhou20223d, han2020reconstructing, xiang2023retro, wen20223d, chen2021unsupervised, jun2023shap}. We mainly focus on learning Neural Implicit Functions with networks for representing 3D shapes or scenes.

\noindent\textbf{Neural Implicit Functions.}
Recently, Neural Implicit Functions (NIFs) have shown promising results in surface reconstruction \cite{park2019deepsdf, ma2021neural, mescheder2019occupancy, chen2023grid, han2020drwr, chen2022latent, jin2023multi, ma2023towards, BaoruiNoise2NoiseMapping}, novel view synthesis \cite{mildenhall2020nerf, barron2021mip, pumarola2021d, zhang2023fast, haque2023instruct, goli2023nerf2nerf, muller2022instant}, image super-resolution \cite{sitzmann2020implicit, muller2022instant, brooks2023instructpix2pix}, point normal estimation \cite{li2022neaf}, etc. In the task of surface reconstruction, the NIFs approaches train a neural network to represent shapes and scenes with binary occupancy \cite{mescheder2019occupancy, peng2020convolutional, chen2019learning} or signed distance functions (SDFs) \cite{park2019deepsdf, jiang2020local, duan2020curriculum, chougensdf, peng2021shape, boulch2022poco, yifan2021iso, ben2022digs}, and then extract surfaces from the learned implicit functions with the marching cubes algorithm \cite{lorensen1987marching}. Previous methods \cite{mescheder2019occupancy, park2019deepsdf} learn a global latent code for each 3D shape with a shared MLP to decode geometries. Some advanced approaches \cite{peng2020convolutional, jiang2020local} propose to use more latent codes to represent detailed local geometries. Latest studies bring more priors learned from large scale datasets to enhance the representation ability of NIFs. PCP \cite{PredictiveContextPriors} introduces the predictive context priors to represent large-scale point clouds. OnSurf \cite{On-SurfacePriors} explores the on-surface prior to reconstruct smooth surfaces from sparse point clouds. 

NeRF \cite{mildenhall2020nerf} and the following studies \cite{muller2022instant, park2021nerfies, tancik2022block, meng2023neat, han2023iccv, rosu2023permutosdf, hertz2022prompt, li2023neuralangelo} extends NIFs to the task of novel view synthesis. NeuS \cite{wang2021neus}, VolSDF \cite{yariv2021volume} and Unisurf \cite{oechsle2021unisurf} advance the rendering procedure of NeRF for surface reconstruction from multi-view images. ManhattanSDF \cite{guo2022neural} leverages the manhattan assumption for representing indoor scenes. NeuRIS \cite{wang2022neuris} and MonoSDF \cite{yumonosdf} brings the depth prior and normal prior for improving the surface quality.

Occupancy and SDFs are merely suitable to represent closed shapes. Recently, researchers explore the neural unsigned distances (UDFs) \cite{chibane2020neural, Zhou2022CAP-UDF, zhao2021learning, venkatesh2021deep, chen20223psdf, wanghsdf, long2022neuraludf, liu2023neudf} to represent shapes with arbitrary topology. NDF \cite{chibane2020neural} trains a neural network to learn UDFs with the ground truth distance values. GIFS \cite{ye2022gifs} jointly learns UDFs and the relationship of queries. MeshUDF \cite{guillard2021meshudf} propose a differentiable meshing algorithm for UDF.  However, all the previous UDF-based methods fail to reconstruct smooth and continuous surfaces with high-fidelity details due to the disability of learning an accurate and smooth zero level set of UDF.

\noindent\textbf{Implicit Learning from Raw Point Clouds.}
With the ground truth signed/unsigned distances or occupancy labels as supervision, previous methods learn NIFs as a regression \cite{chibane2020neural, park2019deepsdf} problem or classification \cite{mescheder2019occupancy} problem. However, these expensive supervisions come from the ground truth meshes which are hard to collect. Moreover, such methods also fail to generalize to unseen cases with large geometry variations to the training data, which limits the application scenarios of such methods. 
To learn implicit representations directly from raw point clouds without ground truth supervisions, current methods explore the supervision from sign agnostic \cite{atzmon2020sal, atzmon2020sald} learning with gradient constraints \cite{gropp2020implicit}. Neural-Pull \cite{ma2021neural} employs a new learning schema to pull space onto surfaces for optimizing SDFs. 

Latest work CAP-UDF \cite{Zhou2022CAP-UDF} learns UDF progressively with a field consistency loss. However, the UDF inferred by CAP-UDF are not accurate and the gradients are unreliable, especially in the zero level set which indicates the location of 3D geometries. 
The key problem is that the differential neural networks struggle from learning non-differentiable zero level set of UDF, leading to highly fragmented and discontinuous surfaces. 
Since the non-zero level sets of UDF are all differentiable, we propose to pull the non-zero level sets onto the zero level set with gradient constraints which align gradients and correct unsigned distance errors on the zero level set, leading to a smoother and more continuous zero level set of unsigned distance field. Our learning schema involves the constraints for both zero level set and non-zero level sets of UDF, which differentiates our method from the previous ones \cite{Zhou2022CAP-UDF,atzmon2020sal, atzmon2020sald,ma2021neural,chibane2020neural, park2019deepsdf}.

\begin{figure*}[!t]
    \centering
    \includegraphics[width=\textwidth]{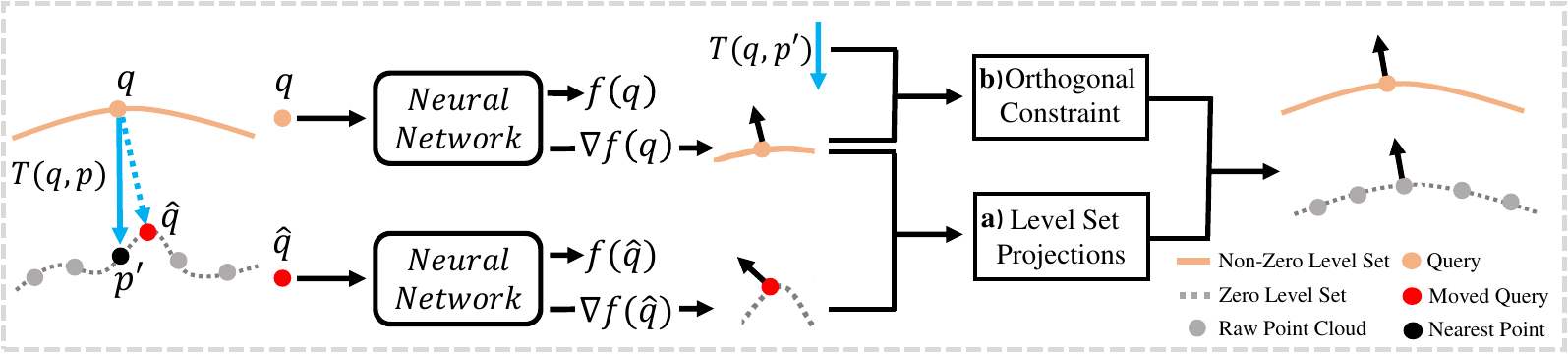}
    \caption{Overview of our method. (a) The level set projection constraint leverages the gradient $\nabla f(q)$ at convincing non-zero level set as the guidance to gradient $\nabla f(\hat{q})$ at unreliable zero level set. (b) The gradient-surface orthogonal constraint encourages the gradients $\nabla f(q)$ at query $q$ orthogonal to the tangent plane of the surface at $p'$ which is the nearest point of $q$ in the raw point cloud.}
    \label{fig:overview}
    \vspace{-0.5cm}
\end{figure*}

\section{Method}

\noindent\textbf{Neural UDFs and Level Sets.} We focus on learning unsigned distance functions for representing 3D shapes or scenes with arbitrary typology. 
Given a raw point cloud $P = \{p_i \in R^3\}_{i=1}^N$ with $N$ points, we sample a set of queries $Q = \{q_j \in R^3\}_{j=1}^M$ around $P$. The UDF $f$  predicts the unsigned distances $s^P$ between queries $\{q_j \}_{j=1}^M$ and the shape described by $P$, formulated as:
\begin{equation}
    s_j^P = f(q_j).
\end{equation}

A level set $S_d$ of UDF is defined as a region of the 3D space containing a continuous set of 3D points with the same unsigned distance $d$, formulates as:

\begin{equation}
    S_d = \{q|f(q)=d\},
\end{equation}
and the level sets of UDF are defined as $\mathcal{S} = \{S_d\}_{d>=0}$.

\noindent\textbf{Learning UDFs from Point Clouds.} 
Inspired by CAP-UDF \cite{Zhou2022CAP-UDF}, we leverage the pulling optimization for learning UDFs from raw point clouds without ground truth unsigned distances as supervision. Specifically, given a query $q_i$ as input, we move it against the direction of the gradient $g_i = \nabla f(q_i)$ at $q_i$, the moving operation is formulated as:
\begin{equation}
    \label{eq:move}
    \hat{q}_i = q_i - f(q_i) \cdot \nabla f(q_i)/||\nabla f(q_i)||_2,
\end{equation}
where the chamfer distance between moved queries $\hat{Q} = \{\hat{q}\}_{i=1}^M$ and the raw point cloud $P$ is calculated as the loss function for optimization, formulated as:

\begin{equation}\small
    \label{eq:gcloss}
    \begin{split}
         \mathcal{L}_{\rm CD} = \frac{1}{M} \sum_{i \in [1,M]}  &\min\limits_{j \in [1, N]}{||\hat{q}_i-p_j||_2} 
        \\&+ \frac{1}{N} \sum_{j \in [1,N]} \mathop{min}\limits_{i \in [1, M]}{||p_j-z_i||_2},
    \end{split}
\end{equation}

\begin{figure}[!t]
    \centering
    \includegraphics[width=1\columnwidth]{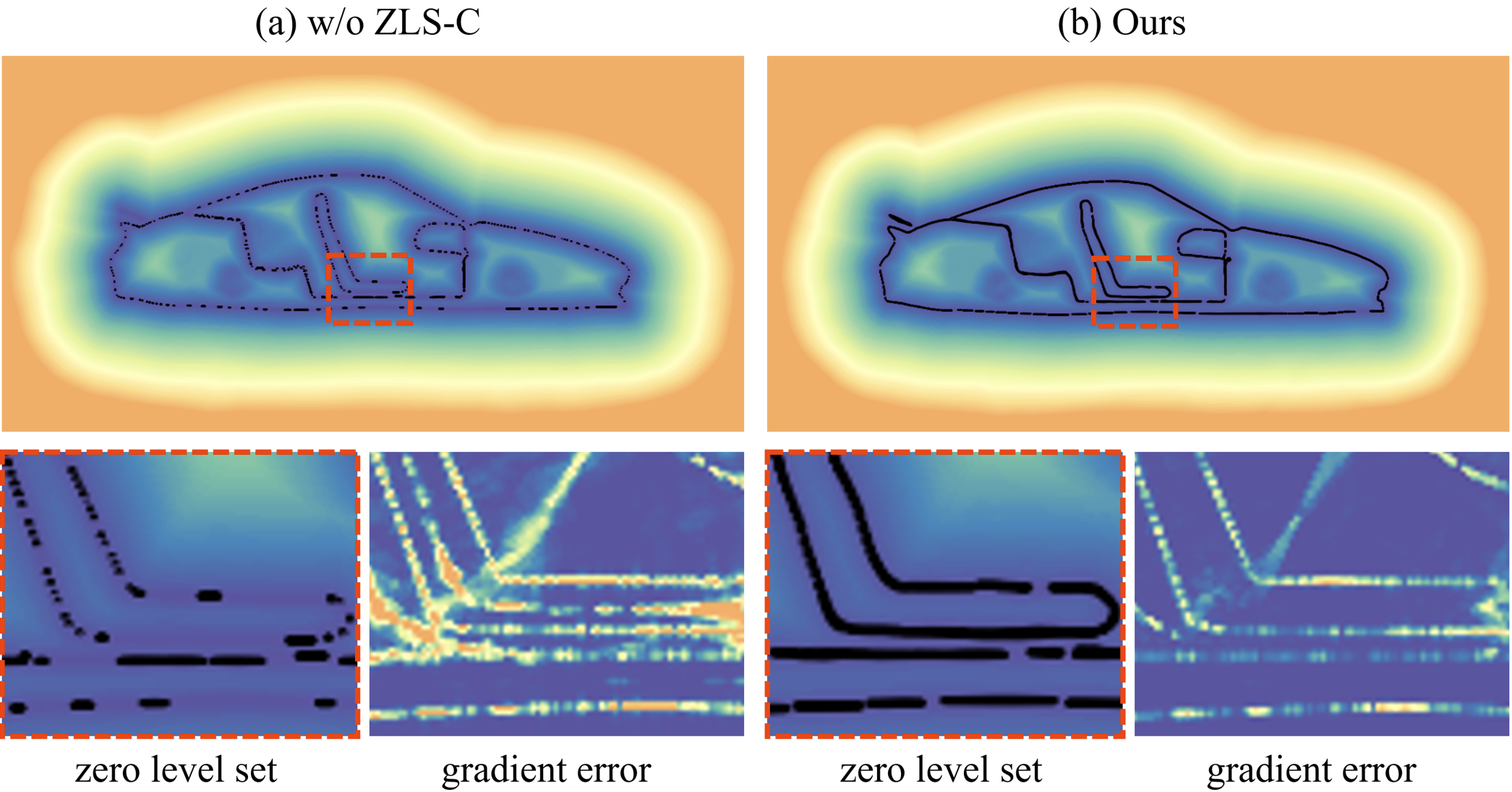}
    \caption{We visualize the fields and the zero level set of UDF learned 
with or without our designed constraints on the zero level set (ZLS-C). The bluer the color, the smaller the unsigned distances in the field and the errors on gradients.}
    \label{fig:zls_error}
    \vspace{-0.5cm}
\end{figure}

\noindent\textbf{Zero Level Set of UDF.}
An accurate zero level set is the key factor for UDF to represent 3D shapes or scenes, since the zero level set indicates the location of 3D geometries. The key problem here is that the differential networks struggle from learning the zero level set where the UDF is not differentiable, which leads to large errors on unsigned distances and gradients around the zero level set.

We provide a visualization of fields and zero level set of UDF learned with neural networks in Fig. \ref{fig:zls_error}. As shown in Fig. \ref{fig:zls_error} (a), the zero level set of UDF learned without proper constraints is highly fragmentary and discontinuous with large errors on the gradients at the space near to the zero level set.  On the contrary, we achieve the accurate and continuous zero level set as shown in Fig. \ref{fig:zls_error} (b) by introducing level set projection constraint and gradient-surface orthogonal constraint to the zero level set.

\begin{figure*}[h]
    \centering
    
    \includegraphics[width=\textwidth]{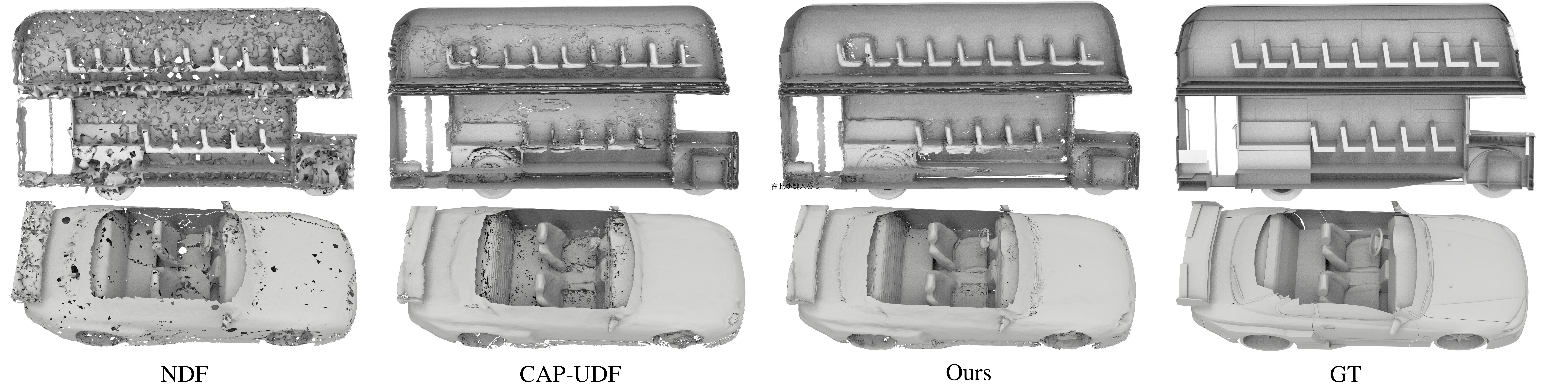}
    \vspace{-0.6cm}
    
    \caption{Visual comparisons of surface reconstruction on ShapeNet dataset.}
    \label{fig:cars}
    \vspace{-0.5cm}
\end{figure*}

\noindent\textbf{Level Set Projections. }
For learning accurate zero level set of UDF, we design level set projection constraint to guide the learning of zero level set using the rest non-zero level sets. Our idea comes from the observation that despite the unreliable zero level set, the other level sets of the learned UDF are quite convincing, which can serve as the guidance to the field optimization on the zero level set. As shown in Fig. \ref{fig:zls_error}, the gradient errors on the non-zero level set are much less than the errors on the zero level set.

We illustrate our method in Fig. \ref{fig:overview}, where we pull the non-zero level set onto the zero level set with gradient constraints which align gradients and correct unsigned distance errors on the zero level set. We achieve this by encouraging the parallelism between the gradient at queries $Q = \{q_i\}_{i=1}^{M}$ and the gradient at their projections $\hat{Q} = \{\hat{q}\}_{i=1}^{M}$ to the zero level set, formulated as:
\begin{equation}
    \mathcal{L}_{\rm proj} = \frac{1}{M} \sum_{i=1}^M{\gamma_i ( 1- \big| \frac{\nabla f(q_i) \cdot \nabla f(\hat{q}_i)}{||\nabla f(q_i)||_2 \cdot ||\nabla f(\hat{q_i})||_2} \big| )}, 
\end{equation}
where $\gamma_i$ is the factor for weighting the gradient constraint at the query $q_i$, indicating the importance of optimizing query $q_i$, formulated as:

\begin{equation}
    \gamma_i = \exp(-\lambda |f(q_i)|).
\end{equation}

We set $\gamma$ to increase when queries are closer to the zero level set where $\lambda$ is a hyper-parameter to control the increasing speed. The nonuniform weighting encouraging the network to focus more on the optimizations near the zero level set of UDF.

While $\mathcal{L}_{proj}$ can provide an effective guidance to the gradients at zero level set, we further explore a more accurate unsigned distance constraint for correcting the distance values at the zero level set of UDF. The implemented loss is defined as:  
\begin{equation}
    \mathcal{L}_{\rm dist} = \frac{1}{N} \sum_{i=1}^N{|f(p_i)|},
\end{equation}
where $p_i \in P$ is the raw point cloud indicating the exact surface to be represented.

\begin{table}[!tb]
\vspace{0.2cm}
    \centering
\resizebox{\linewidth}{!}{
\begin{tabular}{l|cc|cc|c}
\toprule
\multirow{1}*{} & \multicolumn{2}{c}{Chamfer-L2} & \multicolumn{2}{c}{F-Score} & NC\\

Method  &Mean  &Median & $F1^{0.005}$ & $F1^{0.01}$\\ 
\midrule
Input & 0.363 & 0.355 &48.50 &88.34 & - \\
Watertight GT & 2.628 & 2.293 & 68.82 & 81.60 & -\\
GT & 0.076 & 0.074 & 95.70 & 99.99 & -\\
NDF \cite{chibane2020neural} & 0.202 & 0.193 &77.40 & 97.97 & 79.1\\ 
GIFS \cite{ye2022gifs} & 0.128 & 0.123 & 88.05  & 99.31 & -\\
CAP-UDF$_{BPA}$ & 0.141 & 0.138 & 84.84 & 99.33 & 81.8\\
CAP-UDF \cite{Zhou2022CAP-UDF} & 0.119 & 0.114 & 88.55 & 99.82 & 82.5 \\
\midrule

Ours & \textbf{0.098} & \textbf{0.097} & \textbf{92.18} & \textbf{99.90} &\textbf{85.0}\\
\bottomrule
\end{tabular}}
\vspace{0.05cm}
    \caption{Surface reconstruction for point cloud on ShapeNet cars dataset (Chamfer-L2$\times 10^4$).}
    \label{table:cars}
    \vspace{-0.3cm}
\end{table}

\noindent\textbf{Gradient-Surface Orthogonal Constraint. }
The key idea to guide the learning of zero level set using the rest level sets is based on the observation that the non-zero level sets are much more reliable. To bring a more robust guidance to the optimization of zero level set, we introduce an addition gradient constraint on queries to learn a more stable and regular level sets at the non-zero level sets of UDF by enforcing the gradient orthogonal to the tangent plane of the surface. We design a novel loss to encourage the consistency between the gradient direction at a query $q_i$ and the direction from $q_i$ to its closest point $p_i$ on $P$, formulated as:
\begin{equation}
    \mathcal{L}_{\rm orth} = \frac{1}{M} \sum_{i=1}^M{( 1- \big| \frac{\nabla f(q_i) \cdot T(q_i, p_i)}{||\nabla f(q_i)||_2 \cdot ||T(q_i, p_i)||_2} \big| )},
\end{equation}
where $T(q_i, p_i)$ is the direction from $q_i$ to $p_i$. The constraint is illustrated in Fig. \ref{fig:overview}.

A more naive way of introducing the direction vector $T(q_i, p_i)$ for learning is to explicitly substitute the gradient $\nabla f(q_i)$ with the direction vector $T(q_i, p_i)$, as proposed in GenSDF \cite{chougensdf}. However, $T(q_i, p_i)$ is obviously not the exact direction for $q_i$ to reach the approximated surface since the raw point cloud is the discrete representation of the surface, and $p_i$ is not the exact nearest points of $q_i$ in the approximated surface, which leads to wrong supervisions for GenSDF.

On the contrary, We introduce a more flexible way by treating direction vector $T(q_i, p_i)$ as an optimization objective, which enables the networks to accommodate the inconsistency between the closest point $p_i$ on $P$ and the closest point $\hat{q}_i$ predicted by the networks.

\noindent\textbf{Loss Function.} Finally, our loss function is formulated as:
\begin{equation}
    \mathcal{L} = \mathcal{L}_{\rm CD} + \alpha_1 \mathcal{L}_{\rm proj} + \alpha_2 \mathcal{L}_{\rm dist}
    + \alpha_3 \mathcal{L}_{\rm orth},
\end{equation}
where $\alpha_1$, $\alpha_2$ and $\alpha_3$ are the balance weights for our designed three losses to learn accurate zero level set of UDF.

\begin{figure*}[!t]
    \centering
    \includegraphics[width=\textwidth]{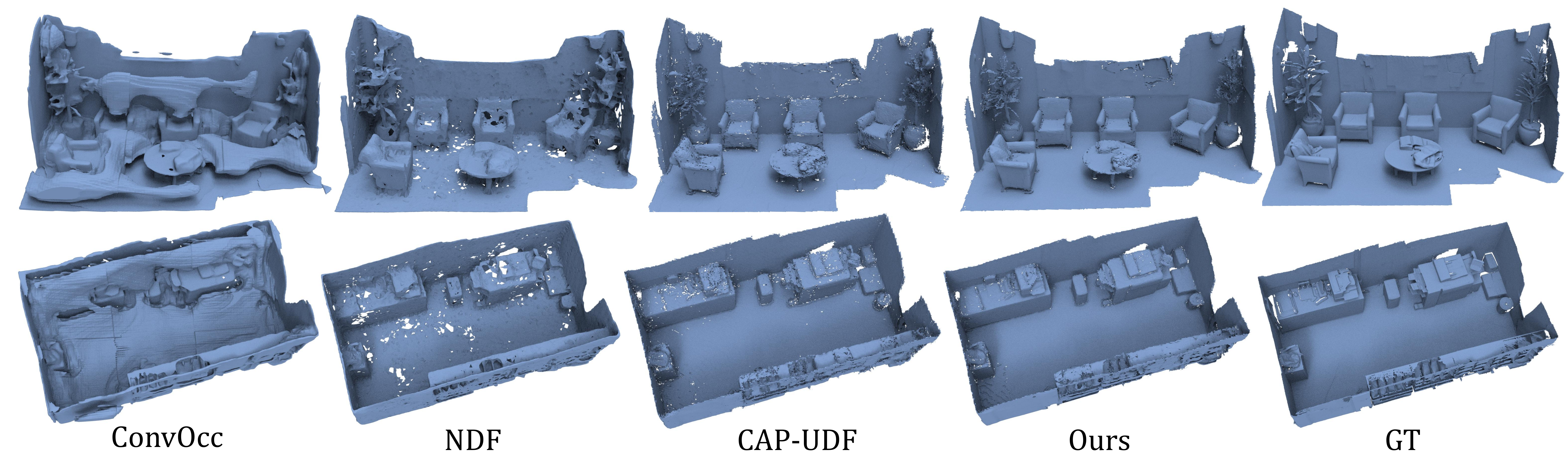}
    \caption{Visual comparisons of surface reconstruction on 3D Scene dataset. Input contains 1K points / $m^2$}
    \label{fig:3dscene}
    \vspace{-0.2cm}
\end{figure*}

\begin{table*}[!]
\centering
\resizebox{\linewidth}{!}{
    \begin{tabular}{c|c|c|c|c||c|c|c||c|c|c||c|c|c||c|c|c}
     \hline

        &&\multicolumn{3}{c||}{Burghers}&\multicolumn{3}{c||}{Lounge}&\multicolumn{3}{c||}{Copyroom}&\multicolumn{3}{c||}{Stonewall}&\multicolumn{3}{c}{Totempole}\\
        \hline
        &&L2CD&L1CD&NC&L2CD&L1CD&NC&L2CD&L1CD&NC&L2CD&L1CD&NC&L2CD&L1CD&NC\\
     \hline

     \hline
     \multirow{7}{*}{\rotatebox{90}{500/$m^2$}}&ConvONet~\cite{peng2020convolutional}&26.97&0.081&0.905&9.044 &0.046 &0.894 &10.08 &0.046 &0.885 &17.70 &0.063 &0.909 &2.165 &0.024 &0.937\\
     &LIG~\cite{jiang2020local}&3.080 &0.046 &0.840 &6.729 &0.052 &0.831 &4.058 &0.038 &0.810&4.919 &0.043 &0.878 &9.38 &0.062 &0.887\\
     &DeepLS~\cite{chabra2020deep}&0.714&0.020&0.923&10.88&0.077&0.814&0.552&0.015&0.907&0.673&0.018&0.951&21.15&0.122&0.927\\
     
     &NDF \cite{chibane2020neural}&0.546&0.018&0.917&0.314&0.012&0.921&0.242&0.012& 0.907 &0.226&0.012&0.949 &1.049&0.025 & 0.939\\
     &OnSurf \cite{On-SurfacePriors}&0.609&0.018&\textbf{0.930}&0.529&0.013&0.926&0.483&0.014&0.908&0.666&0.013&0.955&2.025&0.041&0.954\\

     &CAP-UDF \cite{Zhou2022CAP-UDF} &{0.192}&{0.011}&0.911&{0.099}&{0.009}&0.911&{0.120}&{0.009}&0.902&{0.069}&{0.008}&0.958&{0.131}&{0.010}&0.954\\
     
     \cline{2-17}
     &Ours & \textbf{0.161} & \textbf{0.009 }& 0.916 & \textbf{0.087} & \textbf{0.006} & \textbf{0.933} & \textbf{0.102} & \textbf{0.007} & \textbf{0.920} & \textbf{0.061} & \textbf{0.007} & \textbf{0.964} & \textbf{0.114} & \textbf{0.007} & \textbf{0.960}\\
     
     \hline
     \hline
     \multirow{7}{*}{\rotatebox{90}{1000/$m^2$}}&ConvONet~\cite{peng2020convolutional}&27.46&0.079&0.907&9.54 &0.046 &0.894 &10.97 &0.045 &0.892 &20.46 &0.069 &0.905 &2.054 &0.021 &0.943\\
     &LIG~\cite{jiang2020local}&3.055 &0.045 &0.835 &9.672 &0.056 &0.833 &3.61 &0.036 &0.810&5.032 &0.042 &0.879 &9.58 &0.062 &0.887\\
    &DeepLS~\cite{chabra2020deep}&0.401&0.017&0.920&6.103&0.053&0.848&0.609&0.021&0.901&0.320&0.015&0.954&0.601&0.017&0.950\\
     &NDF \cite{chibane2020neural}&1.168&0.027&0.901&0.393&0.014&0.910&0.269&0.013&0.908&0.509&0.019&0.936&2.020&0.036&0.922\\
     &OnSurf \cite{On-SurfacePriors}&1.339&0.031&\textbf{0.929}&0.432&0.014&0.934&0.405&0.014&0.914&0.266&0.014&0.957&1.089&0.029&0.954\\

     &CAP-UDF \cite{Zhou2022CAP-UDF}&{0.191}&{0.010}&0.910&{0.092}&{0.008}&0.927&{0.113}&{0.009}&0.911&{0.066}&{0.007}&0.962&{0.139}&{0.009}&0.955\\
     
    \cline{2-17}
    &Ours & \textbf{0.146} & \textbf{0.009} & 0.921 & \textbf{0.068} & \textbf{0.006} & \textbf{0.941} & \textbf{0.093} & \textbf{0.007 }& \textbf{0.925} & \textbf{0.050} & \textbf{0.005} & \textbf{0.970} & \textbf{0.107} & \textbf{0.007} & \textbf{0.960}\\
   \hline
   \end{tabular}}
   \vspace{0.02cm}
   \caption{Surface reconstruction for point clouds under 3D Scene. L2CD$\times 1000$.}
   \label{table:t12100}
   \vspace{-0.5cm}
\end{table*}

\noindent\textbf{Applications.} The zero level set of UDF indicates the location of represented 3D geometries. Thus, a well-learned UDF with accurate and continuous zero level set can be adopt to different 3D applications.

\textit{\textbf{Surface Reconstruction}}. Following previous methods \cite{Zhou2022CAP-UDF, guillard2021meshudf}, the surfaces can be extracted from the zero level set of UDF by judging gradient directions, formulated as:
\begin{equation}
    \mathcal{D}(q_i, q_j) = \nabla f(q_i) \cdot \nabla f(q_j),
\end{equation}
to classify whether two queries $q_1$ and $q_2$ are in the same $\mathcal{D}(q_i, q_j) > 0$ or the opposite side $\mathcal{D}(q_i, q_j) < 0$ of the approximated surface. The marching cubes-like algorithm is then used to extract surfaces.

\textit{\textbf{Unsupervised Point Normal Estimation}}. With a well learned zero level set, we compute the gradients at the raw point clouds $P$ as their estimated normals, formulated as:
\begin{equation}
    \mathcal{N}(P) = \{\nabla f(p)\},p\in P.
\end{equation}

\textit{\textbf{Unsupervised Point Cloud Upsamping}}. We randomly sample queries around $P$, and then move them to the zero level set with Eq. \ref{eq:move} as the upsampled points of $P$. We further filter the queries with unsigned distance values larger than a threshold $\beta$ for avoiding outliers. The procedure for upsampling $P$ with a factor of $G$ is formulated as: 

\begin{equation}
    \mathcal{U}(P) = \{\hat{q}_i, i \in \left [ 1, N\times G \right ] | f(q_i) < \beta \}
\end{equation}

\section{Experiments}
For evaluating the performance of our method, we conduct experiments on surface reconstruction from raw point clouds of shapes (Sec. \ref{gen}), room-level scenes (Sec. \ref{scene}), and real scans (Sec. \ref{scan}), where we also show our results on large-scale driving scenes obtained by fusing LiDAR depths in Sec. \ref{kitti}. We further explore the ability of our learned UDF with accurate zero level set for unsupervised point normal estimation (Sec. \ref{normal}) and unsupervised point upsampling (Sec. \ref{upsamp}). Finally, we conduct ablation studies in Sec. \ref{ablation}.

\begin{figure*}[!t]
    \centering
    \includegraphics[width=\textwidth]{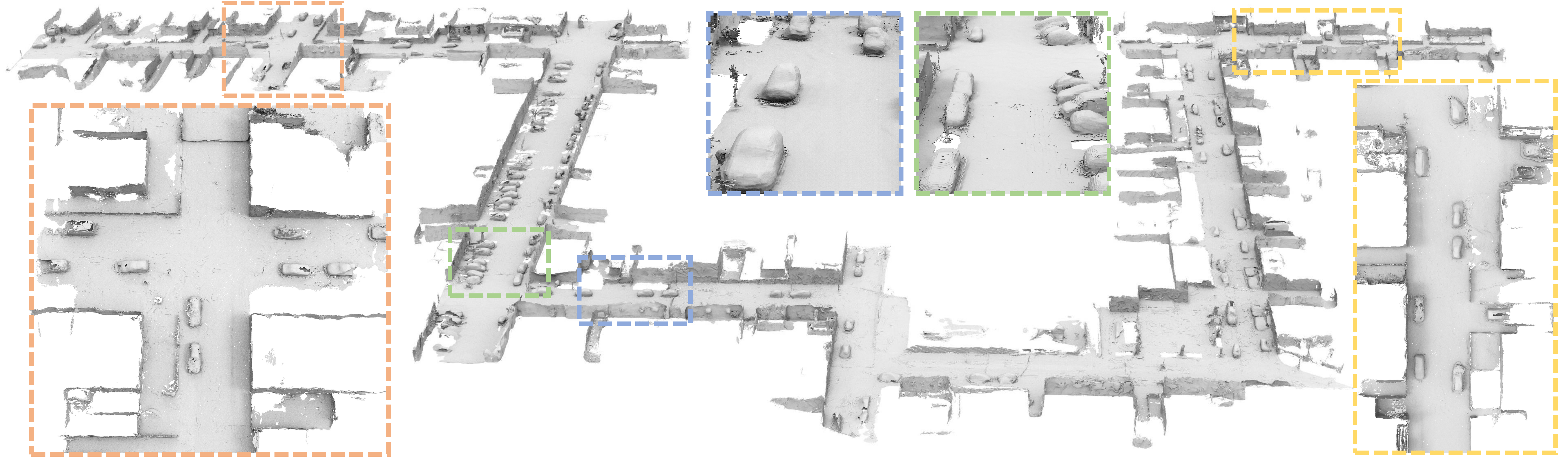}
    \caption{Surface reconstruction results on KITTI odometry dataset (Sequence 00, frame 3000 to 4000) .}
    \label{fig:kitti}
    \vspace{-0.3cm}
\end{figure*}

\subsection{Surface Reconstruction  for Shapes}

\label{gen}
\noindent\textbf{Dataset and Metrics.} To explore the ability of our method in representing shapes with arbitrary topology, we follow previous methods \cite{Zhou2022CAP-UDF, ye2022gifs} to conduct experiments on the ``Car'' category of ShapeNet dataset which contains the largest number of shapes with multi-layer structures or open surfaces. We sample 10K points per shape as the input for surface reconstruction. For evaluating the performances, we follow the common setting \cite{ye2022gifs} to sample 100K points from the reconstructed surfaces and leverage Chamfer Distance, Normal Consistency \cite{mescheder2019occupancy} and F-score with a threshold of 0.005 / 0.01 as the evaluation metrics.

\noindent\textbf{Comparisons.}
We compare our method with the state-of-the-art shape reconstruction methods NDF \cite{chibane2020neural}, GIFS \cite{ye2022gifs} and CAP-UDF \cite{Zhou2022CAP-UDF}. We report the quantitative comparison results in Tab. \ref{table:cars}, where ``Watertight GT'' is the result of sampling points from the watertight ground truth, indicating the best performance of the methods base on SDF or occupancy. We also show the superior limit of the dataset by sampling points on the ground truth mesh and report the result as ``GT''. Tab. \ref{table:cars} shows that our method achieves the best performance over all the baselines in terms of all metrics. The visual comparison in Fig. \ref{fig:cars} further shows that our methods produce more continuous and accurate geometries over all baselines.

\subsection{Surface Reconstruction for Room-Level Scenes}
\label{scene}
\noindent\textbf{Dataset and Metrics.} We further explore our performance on real scene scans to demonstrate the ability of our methods to scale to large-scale point clouds. Following OnSurf \cite{On-SurfacePriors}, we conduct experiments on the 3D Scene dataset consisting of complex scenes with open surfaces and scan noises. The input point cloud is achieved by sampling 500 or 1000 points per $m^2$ uniformly for each scene. We follow the common setting \cite{On-SurfacePriors, Zhou2022CAP-UDF} to sample 1$M$ points on both the reconstructed and the ground truth meshes. We leverage L1 / L2 Chamfer Distance and Normal Consistency as the evaluation metrics.

\noindent\textbf{Comparisons.}
With different point densities, we compare our method with the state-of-the-art methods ConvONet \cite{peng2020convolutional}, LIG \cite{jiang2020local}, DeepLS \cite{chabra2020deep}, OnSurf \cite{On-SurfacePriors}, NDF \cite{chibane2020neural} and CAP-UDF \cite{Zhou2022CAP-UDF}. The numerical comparison in Tab. \ref{table:t12100} demonstrates our superior performance in all scenes. We further present a visual comparison in Fig. \ref{fig:3dscene}, where the input point clouds contains 1K points/$m^2$. The visualization further shows that our method can reconstruct more detailed and smooth surfaces in complex scenes, where the other UDF-base methods NDF and CAP-UDF produce fragmented surfaces.

\begin{table}[h]\scriptsize
    \centering
    \setlength{\tabcolsep}{2.5mm}
\resizebox{0.95\linewidth}{!}{
    \begin{tabular}{l|c|c|c}
    \toprule
    
    Method  &Chamfer-L1  &F-Score & NC\\ 
    \midrule
    
    IGR \cite{gropp2020implicit} & 0.178 & 75.5 & 86.9\\
    Point2Mesh \cite{hanocka2020point2mesh} & 0.116 & 64.8 & -\\
    SPSR \cite{kazhdan2013screened} & 0.232 & 73.5 & -\\
    SAP \cite{peng2021shape} & 0.076 & 83.0 & 88.6\\
    Neural-Pull \cite{ma2021neural} & 0.106 & 79.7 & 87.2\\
    NDF \cite{chibane2020neural} & 0.238 & 68.6 & 80.4\\
    CAP-UDF \cite{Zhou2022CAP-UDF} & {0.073} & {84.5} & 88.6\\
    
    \midrule
    Ours & \textbf{0.071} & \textbf{85.1} & \textbf{91.0}\\
    \bottomrule

    \end{tabular}}
    \vspace{0.1cm}
    \caption{Surface reconstruction for point cloud on SRB dataset.}
    \label{table:srb}
    \vspace{-0.5cm}
\end{table}

\subsection{Large-Scale Driving Scene Reconstruction}
\label{kitti}
\noindent\textbf{Dataset.} To further demonstrate our scalability to extremely large scenes, we follow NGS \cite{huang2022neural} to apply our method to KITTI dataset \cite{geiger2012we}. We use the LiDAR scans of frame 3000 to 4000 in Squeece00 subset of KITTI dataset as the full input, which are transformed into world coordinates using the provided camera trajectories. We leverage the slide-window strategy to divide the full scene into 15 blocks of size 51.2$m^3$, where we sample 300K points per block as the local input to train our network. The final scene is achieve by blending the reconstructed local scenes. 

\noindent\textbf{Analysis.}
The reconstruction results under KITTI scenes is shown in Fig. \ref{fig:kitti}, where our method reproduces visual-appealing performances with accurate geometries. Note that: 1) The LiDAR scans contain large amount of noises, especially in the difficult driving scenes. 2) Our method learns to reconstruct surfaces directly from the raw point cloud without learning priors from large-scale datasets.

\subsection{Surface Reconstruction for Real Scans}
\label{scan}
\noindent\textbf{Dataset and Metrics.}
We demonstrate our advantages for surface reconstruction from real scans by conducting experiments on the Surface Reconstruction Benchmarks \cite{williams2019deep} (SRB). We follow previous methods \cite{peng2021shape, Zhou2022CAP-UDF} to leverage Chamfer Distance and F-Score with a threshold of 1$\%$ for evaluation. Note that the ground truth of SRB dataset is dense point cloud containing millions of points. For a comprehensive comparison on the reconstruction quality, we further evaluate the Normal Consistency by first converting the ground truth into meshes with BPA \cite{bernardini1999ball} algorithm.

\begin{figure}[!t]
    \centering
    \includegraphics[width=\columnwidth]{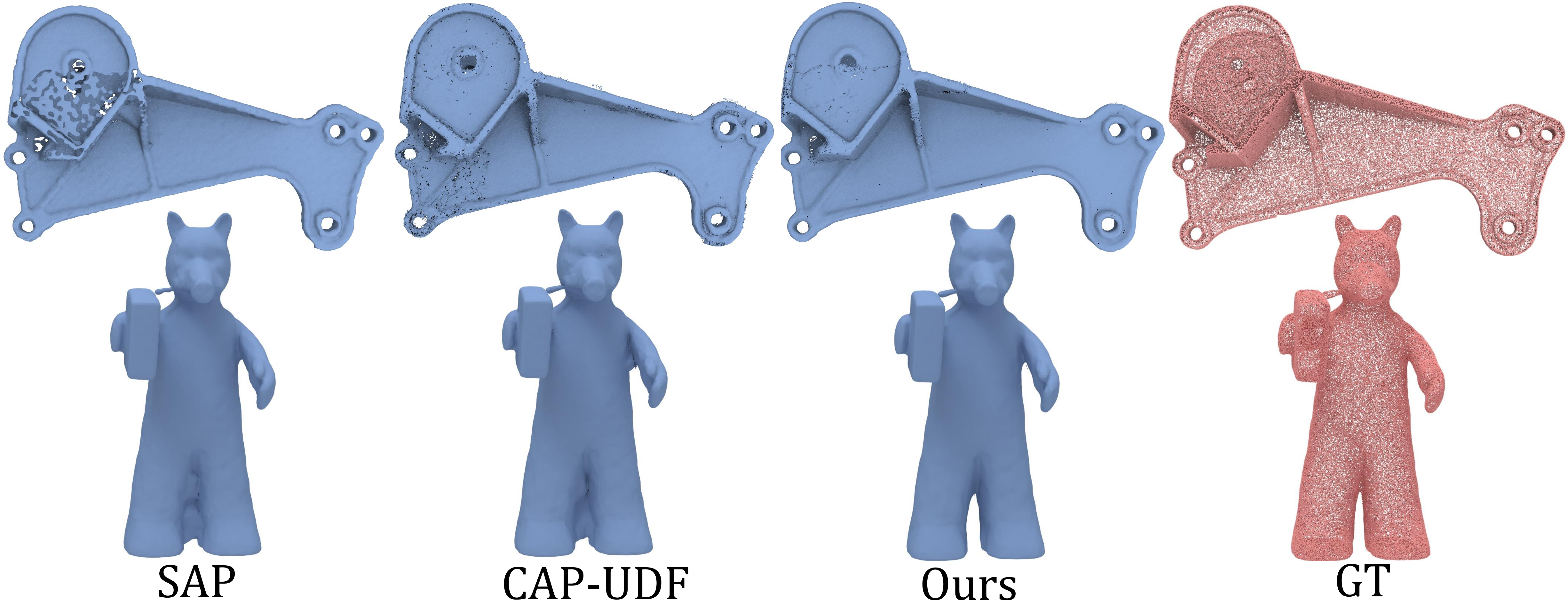}
    \caption{Visual comparisons of surface reconstruction on SRB dataset. SRB dataset provides only point clouds as GT.}
    \label{fig:srb}
\end{figure}

\noindent\textbf{Comparisons.}
We first report our numerical comparison under SRB dataset with the classic and learning-based methods in Tab. \ref{table:srb}. The compared methods include IGR \cite{gropp2020implicit}, Point2Mesh \cite{hanocka2020point2mesh}, Screened Poisson Surface Reconstruction (SPSR) \cite{kazhdan2013screened}, Shape As Points (SAP) \cite{peng2021shape}, Neural-Pull \cite{ma2021neural}, NDF \cite{chibane2020neural} and CAP-UDF \cite{Zhou2022CAP-UDF}. Our method achieves the best performance in all metrics as shown in Tab. \ref{table:srb}. The visual comparison in Fig. \ref{fig:srb} further shows that our methods produce more accurate and smooth geometries over all baselines.

\begin{table}\small
\begin{center}
\setlength{\tabcolsep}{1mm}
\resizebox{\linewidth}{!}{
\begin{tabular}{c|l|ccc|c}
\hline\noalign{\smallskip}
~ & Method & Clean & Strip & Gradient & average\\
\noalign{\smallskip}
\hline
\noalign{\smallskip}
\multirow{5}*{Supervised} 
& PCPNet \cite{guerrero2018pcpnet} & 9.66 & 11.47 & 13.42 & 11.61\\
~ & Hough \cite{boulch2016deep} & 10.23 & 12.47 & 11.02 & 11.24\\
~ & Nesti-Net \cite{ben2019nesti} & 6.99 & 8.47 & 9.00 & 8.15\\
~ & IterNet \cite{lenssen2020deep} & 6.72 & 7.73 & 7.51 & 7.32\\
~ & DeepFit \cite{ben2020deepfit} & 6.51 & 7.92 & 7.31 & 7.25\\
\noalign{\smallskip}
\hline
\noalign{\smallskip}
\multirow{4}*{Unsupervised} & 
 Jet \cite{cazals2005estimating} & 12.23 & 13.39 & 13.13 & 12.92\\
~ & PCA \cite{abdi2010principal} & 12.29 & 13.66 & 12.81 & 12.92\\

~ & Ours (w/o ZLS-c) & 7.30 & 7.74 & 7.78 & 7.61 \\
~ & Ours & \textbf{6.51} & \textbf{7.45} & \textbf{7.64} & \textbf{7.20} \\
\noalign{\smallskip}

\hline

\end{tabular}}
\end{center}
\vspace{-0.2cm}
\caption{Point normal estimation results on the PCPNet dataset.}
\label{table:normal}
\vspace{-0.4cm}
\end{table}

\begin{figure}[!t]
    \centering
    \includegraphics[width=1\columnwidth]{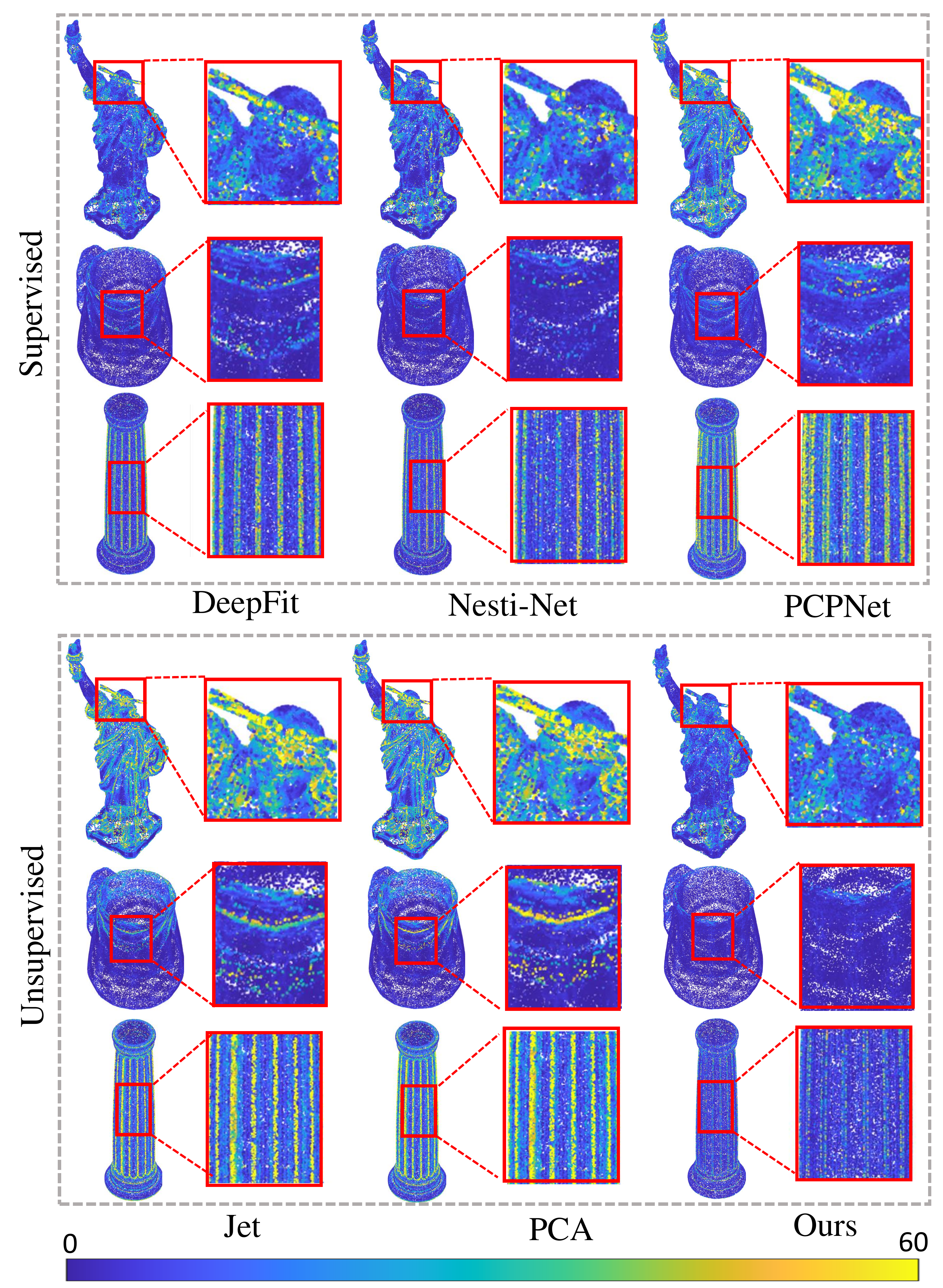}
    \caption{Visual comparisons on error maps of point normal estimation on PCPNet dataset.}
    \label{fig:normal}
\end{figure}

\subsection{Unsupervised Point Normal Estimation}
\label{normal}

\noindent\textbf{Dataset and Metrics.}
We adopt the widely-used PCPNet \cite{guerrero2018pcpnet} dataset to evaluate our method on the task of point normal estimation. Each point cloud in the PCPNet dataset contains 100K points. We conduct experiments under three different settings. The inputs of ``Clean'' setting are sampled uniformly on shapes, and two additional settings with varying point densities (Stripes and Gradients) are added to explore our performance in handling irregular data.  Following previous methods \cite{ben2020deepfit, guerrero2018pcpnet}, we sample 5000 points per shape and compute the angle root mean square error (RMSE) between the predicted normals and the ground truth normals as the evaluation metrics.  

\noindent\textbf{Comparisons.}
We conduct a comparison with traditional normal estimation methods including PCA \cite{abdi2010principal}, Jets \cite{cazals2005estimating} and the state-of-the-art learning-based methods PCPNet \cite{guerrero2018pcpnet}, HoughCNN \cite{boulch2016deep}, Nesti-Net \cite{ben2019nesti}, Iter-Net \cite{lenssen2020deep} and DeepFit \cite{ben2020deepfit}. Note that all the previous learning-based methods are designed in a supervised manner which requires a large scale dataset for training, while we extract normals directly from the unsigned distance fields learned from raw point cloud in an unsupervised way. The numerical results in Tab. \ref{table:normal} show that our method achieves a comparable or even better performance than the state-of-the-art supervised methods. The results also show the effectiveness of our method where the performance degrades from $7.20$ to $7.61$ without our constraints. The visual comparison in Fig. \ref{fig:normal} shows that our method produce more accurate estimations compared to other methods, especially on the complex geometries.

\begin{table}\small
\begin{center}
\setlength{\tabcolsep}{2.5mm}
\resizebox{\linewidth}{!}{
\begin{tabular}{l|c|ccc}
\hline\noalign{\smallskip}
 Method & Supervised? &P2F & CD & HD \\
\noalign{\smallskip}
\hline
\noalign{\smallskip}

PU-Net \cite{yu2018pu} & Yes & 6.84 & 0.72 & 8.94\\
MPU \cite{yifan2019patch} & Yes & 3.96 & 0.49 & 6.11\\
PU-GAN \cite{li2019pu} & Yes & 2.33 & 0.28 & 4.64\\
Dis-PU \cite{li2021point} & Yes & 2.01 & 0.22 & 2.83\\
\noalign{\smallskip}
\hline
\noalign{\smallskip}

EAR \cite{huang2013edge}  & No & 5.82 & 0.52 & 7.37\\
L2G-AE \cite{liu2019l2g} & No & 39.37 & 6.31 & 63.23\\
SPU-Net (Tr2T) \cite{liu2022spu}& No  & 5.97 & 0.38 & 2.24 \\
SPU-Net (All2T)& No  & 5.79 & 0.37 & 2.55 \\
SPU-Net (Te2T)& No  & 6.85 & 0.41 & 2.18 \\
\noalign{\smallskip}
\hline
\noalign{\smallskip}
Ours (w/o ZLS-C) & No & 1.43  & 0.24 & 2.16 \\
Ours & No & \textbf{1.02}  & \textbf{0.19} & \textbf{1.55} \\
\noalign{\smallskip}

\hline

\end{tabular}}
\end{center}
\vspace{-0.1cm}
\caption{Point cloud upsampling results on the PU-GAN dataset.}
\label{table:upsamp}
\vspace{-0.4cm}
\end{table}

\begin{figure*}[!t]
    \centering
    \includegraphics[width=\textwidth]{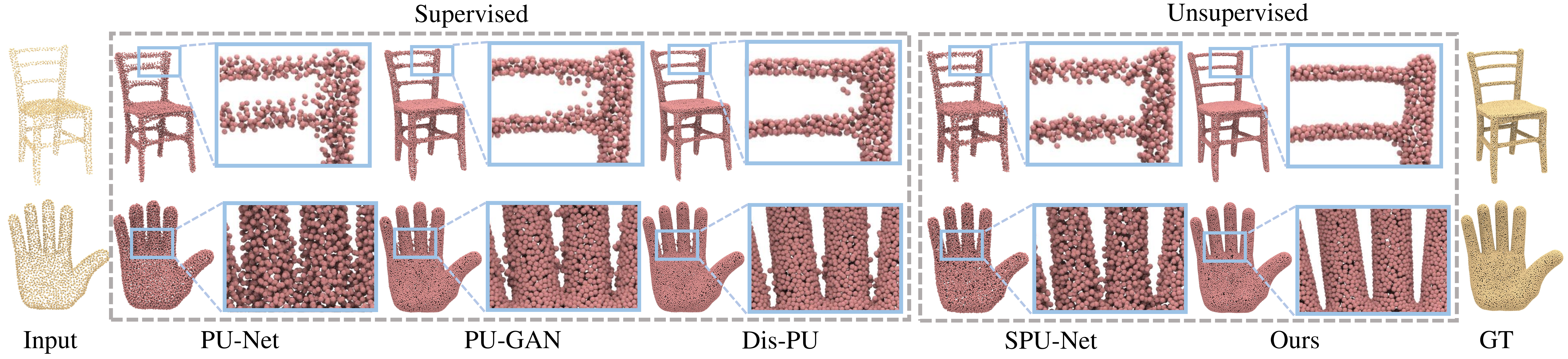}
    \caption{Visual comparisons of point cloud upsampling on PU-GAN dataset.}
    \label{fig:upsamp}
    \vspace{-0.3cm}
\end{figure*}

\subsection{Unsupervised Point Cloud Upsampling}
\label{upsamp}

\noindent\textbf{Dataset and Metrics.}
For the task of point cloud upsampling, we follow previous methods \cite{liu2022spu, li2021point} to conduct experiments on the dataset provided by PU-GAN \cite{li2019pu}. We evaluate our method under the setting to generate 8,198 points for each sparse input with 2,048 points. We adopt Chamfer Distance (CD), Hausdorff Distance (HD) and Point-to-Surface distance (P2F) as the evaluation metrics.

\noindent\textbf{Comparisons.}
We compare our method with the state-of-the-art supervised methods PU-Net \cite{yu2018pu}, MPU \cite{yifan2019patch}, PU-GAN \cite{li2019pu}, Dis-PU \cite{li2021point} and unsupervised methods EAR \cite{huang2013edge}, L2G-AE \cite{liu2019l2g} and SPU-Net \cite{liu2022spu}. The quantitative comparison is shown in Tab. \ref{table:upsamp}, where our method outperforms the previous supervised and unsupervised methods significantly. The results also show the effectiveness of our method where the performance degrades from $1.55$ to $2.16$ in terms of HD without our designed constraints. The visualization in Fig. \ref{fig:upsamp} further shows that our upsampling results are more smooth and uniform even compared with the state-of-the-art supervised methods. 

\subsection{Ablation Studies}
\label{ablation}
We conduct ablation studies to demonstrate the effectiveness of each proposed constraint and explore the effect of some important hyper-parameters. We report the performance under a subset of ShapeNet Cars dataset in terms of L2 Chamfer Distance ($\times 10^4$) and Normal Consistency.  

\noindent \textbf{Effect of each Design.}
We first justify the effectiveness of each proposed constraint in Tab. \ref{tab:ablation1}. We report the performance without all the three designed losses as ``w/o ZLS-C''. We explore each loss by reporting the results without level set projection loss as ``w/o $\mathcal{L}_{\rm proj}$'', surface unsigned distance loss as ``w/o $\mathcal{L}_{\rm dist}$'', gradient-surface orthogonal loss as ``w/o $\mathcal{L}_{\rm orth}$'' and adaptive point weight as ``w/o AW''. We replace our level set projection constraint with the GenSDF-style \cite{chougensdf} optimization as ``GenSDF-Style''. We further constrain the level set projection loss on a randomly selected non-zero level set other than the zero level set and report the performance as ``Non-zero LSC''.
The result demonstrates that the zero level set is the key factor that influence the representation ability of UDF.

\begin{table}[h]\scriptsize
\setlength{\tabcolsep}{0.7mm}
\resizebox{\linewidth}{!}{
 \begin{tabular}{c|cccc}
 \hline
 ~ & w/o ZLS-C & w/o $\mathcal{L}_{\rm proj}$ & w/o $\mathcal{L}_{\rm dist}$ & w/o $\mathcal{L}_{\rm orth}$ \\
 \hline
 L2CD & 0.126  & 0.107 & 0.101 & 0.096\\
 NC & 82.7 & 82.9 & 83.9 & 84.1 \\
 \hline
 \hline
 
 ~ & w/o AW & GenSDF-Style & Non-zero LSC & Ours \\
 \hline
 
 L2CD & 0.096 & 0.101 & 0.098 & \textbf{0.091} \\
 NC & 84.2 & 83.9 & 83.5 &\textbf{ 84.4} \\
 \hline
 
 \end{tabular}}
 \caption{Effect of framework design.}
 \label{tab:ablation1}
 \vspace{-0.1cm}
 
\end{table}

\noindent \textbf{Weight of Level Set Projection Loss.}
We report the performance with different weights of the level set projection loss $\mathcal{L}_{\rm proj}$ in Tab. \ref{tab:ablation2}. We found that a suitable weight of $\mathcal{L}_{\rm proj}$ will improve the reconstruction accuracy, and it may affect the optimization to converge if we weight too much. 

\begin{table}[h]\scriptsize
\setlength{\tabcolsep}{4mm}
\resizebox{\linewidth}{!}{
 \begin{tabular}{c|cccc}
 \hline
 ~ & 0.001 & 0.002 & 0.01 & 0.1 \\
 \hline
 L2CD & 0.095  & \textbf{0.091} & 0.162 & 0.359\\
 NC & 84.2 & \textbf{84.4} & 83.4 & 79.6\\
 \hline
 \end{tabular}}
 \caption{Weight of level set projection loss.}
 \label{tab:ablation2}
 \vspace{-0.1cm}
 
\end{table}

\noindent \textbf{Weight of Surface Unsigned Distance Loss.}
We further explore the effect of weights $\alpha_2$ for the surface unsigned distance loss in Tab \ref{tab:ablation3}, where we found that a too large or too small weight will degenerate the performance.

\begin{table}[h]\scriptsize
\vspace{0.2cm}
\setlength{\tabcolsep}{4mm}
\resizebox{\linewidth}{!}{
 \begin{tabular}{c|cccc}
 \hline
~ & 0.001 & 0.01 & 0.1 & 1 \\
 \hline
 L2CD & 0.101  & 0.100 & \textbf{0.091} & 0.449\\
 NC & 84.1 & 84.1 & \textbf{84.4} & 78.9\\
 \hline
 \end{tabular}}
 \caption{Weight of surface unsigned distance loss.}
 \label{tab:ablation3}
 \vspace{-0.4cm}
 
\end{table}

\noindent \textbf{Weight of Gradient-Surface Orthogonal Loss.}
We evaluate the effect of the weight $\alpha_3$ for gradient-surface orthogonal loss in Tab. \ref{tab:ablation4}. A proper weight $\alpha_3 = 0.01$ brings the best improvement while a too much weight will affect the convergence.

\begin{table}[h]\scriptsize
\setlength{\tabcolsep}{4mm}
\resizebox{\linewidth}{!}{
 \begin{tabular}{c|cccc}
 \hline
 ~ & 0.001 & 0.01 & 0.1 & 1 \\
 \hline
 L2CD & 0.094  & \textbf{0.091} & 0.335 & 2.55\\
 NC & 84.2 & \textbf{84.4} & 77.7 & 72.9\\
 \hline
 \end{tabular}}
 
 \caption{Weight of gradient-surface orthogonal loss.}
 \label{tab:ablation4}
 \vspace{-0.1cm}
\end{table}

\noindent \textbf{Effect of Adaptive per Point Weights.}
We further report the performance under different $\lambda$ which controls the increasing speed of adaptive weights for each point in Tab. \ref{tab:ablation5}. We found that $\lambda$ has a relative small impact on the Chamfer Distance with large affect on the Normal Consistency, which indicates that the adaptive per point weight can improve the surface smoothness. 

\begin{table}[h]\scriptsize
\setlength{\tabcolsep}{4mm}
\resizebox{\linewidth}{!}{
 \begin{tabular}{c|cccc}
 \hline
 ~& 1 & 10 & 100 & 1000 \\
 \hline
 L2CD & 0.093  & \textbf{0.091} & 0.092 & 0.094\\
 NC & 84.1 & \textbf{84.4} & 83.7 & 83.2\\
 \hline
 \end{tabular}}
 \caption{Effect of adaptive per point weights.}
 \label{tab:ablation5}
\end{table}

\section{Conclusion}
In this paper, we justify that an accurate and continuous zero level set of unsigned distance functions is the key factor to represent complex 3D geometries. We therefore propose two novel constraints on UDF to achieve a more continuous zero level set by projecting non-zero level set to guide the optimization of zero level set. With the ability of learning consistent and accurate zero level set of UDF, we explore three different 3D applications including surface reconstruction, unsupervised point normal estimation and unsupervised point cloud upsampling, where the visual and numerical comparisons with the state-of-the-art supervised/unsupervised methods justify our effectiveness ans show our superiority over the latest methods.

\section{Acknowledgement}
This work was supported by National Key R\&D Program of China (2022YFC3800600), the National Natural Science Foundation of China (62272263, 62072268), and in part by Tsinghua-Kuaishou Institute of Future Media Data.

{\small
\bibliographystyle{ieee_fullname}
\bibliography{egbib}
}

\end{document}